\documentclass{svmult}

\usepackage{mathptmx}
\usepackage{helvet}
\usepackage{courier}
\usepackage{amsmath,amsfonts}
\usepackage{graphicx}
\usepackage[bottom]{footmisc}

\begin{document}

\title*{Stiffness modeling of robotic manipulator with gravity compensator}
\author{Alexandr Klimchik$^{a,b}$, Stéphane Caro$^{a,c}$, Yier Wu$^{a,b}$, Damien Chablat$^{a,c}$, 
Benoit Furet$^{a,d}$, Anatol Pashkevich$^{a,b}$}
\institute{%
$^{a}$Institut de Recherche en Communications et Cybernétique de Nantes, France \at
$^{b}$Ecole des Mines de Nantes, France \at
$^{c}$National Center for Scientific Research, France \at
$^{d}$University of Nantes, France \at
\email{alexandr.klimchik@mines-nantes.fr, stephane.caro@irccyn.ec-nantes.fr,  \at
yier.wu@mines-nantes.fr,  damien.chablat@irccyn.ec-nantes.fr, benoit.furet@univ-nantes.fr, \at
anatol.pashkevich@mines-nantes.fr}
}

\titlerunning{Stiffness modeling of robotic manipulator with gravity compensator}
\authorrunning{A. Klimchik et al.}

\maketitle

\abstract{%
  The paper focuses on the stiffness modeling of robotic manipulators with gravity compensators. The main attention is paid to the development of the stiffness model of a spring-based compensator located between sequential links of a serial structure. The derived model allows us to describe the compensator as an equivalent non-linear virtual spring integrated in the corresponding actuated joint. The obtained results have been efficiently applied to the stiffness modeling of a heavy industrial robot of the Kuka family.}

\keywords{Stiffness modeling, gravity compensator, industrial robot.}



\section{Introduction}
Recently, in aerospace industry much attention is paid to the high-precision and high-speed machining of large dimensional aircraft components. To satisfy these requirements, industrial robots are more and more used to replace conventional CNC-machines, which are limited with their performances and are suitable for a rather limited workspace. However, some new problems arise here because of es-sential machining forces caused by processing of modern aeronautic materials that may reduce the quality of technological process. To overcome this difficulty, the robot manufactories attempt to make the manipulators stiffer but quite heavy. The problem of link weights is often solved by using the gravity compensator, which, in its turn, influences on the position accuracy and stiffness properties. The latter motivates enhancement of the manipulator stiffness modeling techniques that are in the focus of this paper. 
The problem of stiffness modeling for serial manipulators has been studied in the robotic community from different aspects [1-3]. In particular, special attention has been paid to heavy industrial and medical robots with essential deflections of the end-effector [4-6]. Among a number of existing stiffness modeling ap-proaches, the Virtual Joint Modeling (VJM) method looks the most attractive for the robotic applications. Its main idea is to take into account the elastostatic prop-erties of flexible components by presenting them as equivalent virtual springs lo-calized in the actuated or passive joints [7]. Because of its simplicity and effi-ciency, this approach has been applied to numerous case-studies and progressively enhanced to take into account specific features of robotic manipulators, such as in-ternal and external loadings, closed-loops, etc. [8, 9]. Besides, some authors have extended the VJM approach by using advanced 6 d.o.f. virtual springs describing stiffness properties of the manipulator elements [10]. However, the problem of the stiffness modeling of the manipulators with gravity compensators has not been studied yet sufficiently; there are only limited number of works that addressed this issue. Besides, in previous works, the main attention has been paid to the compen-sator design [11] and modification of the inverse kinematics algorithms integrated in the robot controller [12].
In contrast to previous works, this paper deals with the stiffness modeling of serial robotic manipulators equipped with a spring-based compensator located be-tween sequential links. It proposes an extension of the VJM-based approach al-lowing to integrate the elasto-static properties of the gravity compensator (that creates a closed loop) into the conventional stiffness model of the manipulator.

\section{Stiffness model of gravity compensator}

Let us consider the general type of gravity compensator that incorporates a passive spring attached between two sequential links of the manipulator (Fig. 1). In a such architecture, there is a closed loop that generates an additional torque in the manipulator joint. So, the specificity of such design allows us to limit modifications of the robot stiffness model by adjusting the virtual joint stiffness parameters.

\begin{figure}[bt]
\center
\includegraphics{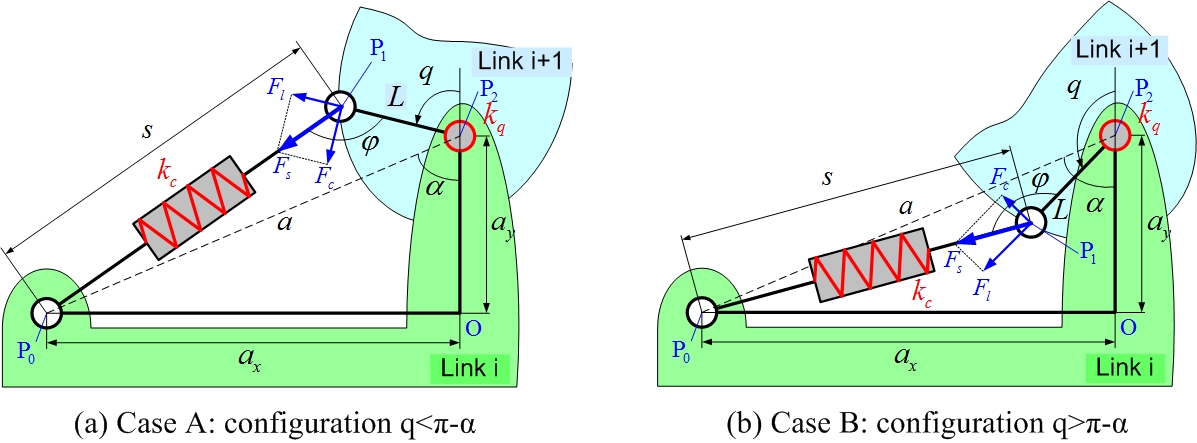}
\caption{Mechanics of gravity compensator.}
\label{Figure:1}
\end{figure}

The geometrical model of considered compensator includes three main node points $P_{0}$, $P_{1}$, $P_{2}$. Let us denote corresponding distances as $L=\left| {{P}_{1}},{{P}_{2}} \right|$, $a=\left| {{P}_{0}},{{P}_{2}} \right|$, $s\,=\left| {{P}_{0}},{{P}_{1}} \right|$. In addition, let us introduce the angles $\alpha $, $\varphi $ and the distances ${{a}_{x}}$ and ${{a}_{y}}$, whose geometrical meaning is clear from Fig. 1. Using these notations, the compensator spring deflection $s$ can be computed as

\begin{equation}\label{Eq:1}
s=\sqrt{{{a}^{2}}+{{L}^{2}}+2\cdot a\cdot L\cdot \cos (\alpha +q)}
\end{equation}
and evidently depends on the joint angle $q$.
The considered mechanical design allows us to balance the manipulator weight for any given configuration by adjusting the compensator spring preloading. It can be taken into account by introducing the zero-value of the compensator length ${{s}_{0}}$ corresponding to the unloaded spring. Under this assumption, the compensator force applied to the node P1 can be expressed as ${{F}_{s}}=K_{c}^{{}}\cdot (s-{{s}_{0}})$, where $K_{c}^{{}}$ is the compensator spring stiffness.
Further, the angle $\varphi$ between the compensator links P0P1 and P1P2 (see Fig. 1) can be found from the expression 	$\sin \varphi ={a}/{s}\;\,\cdot \sin (\alpha +q)$, which allows us to compute the compensator torque ${{M}_{c}}$ applied to the second joint as

\begin{equation}\label{Eq:2}
{{M}_{c}}=K_{c}^{{}}\cdot (1-{{s}_{0}}/s)\cdot a\cdot L\,\cdot \sin (\alpha +q)
\end{equation}
After differentiation of the latter expression with respect to $q$, the equivalent stiffness of the second joint (comprising both the manipulator and compensator stiffnesses) can be expressed as:

\begin{equation}\label{Eq:3}
{{K}_{\theta }}=K_{\theta }^{0}+{{K}_{c}}\cdot \,a\,L\,\cdot \,{{\eta }_{q}}
\end{equation}
where the coefficient

\begin{equation}\label{Eq:4}
{{\eta }_{q}}=\cos (\alpha +q)-\frac{{{s}_{0}}}{s}\,\cdot \left( \frac{a\,L}{{{s}^{2}}}{{\sin }^{2}}(\alpha +q)+\cos (\alpha +q) \right)
\end{equation}
highly depends on the value of joint variable $q$ and the initial preloading in the compensator spring described by ${{s}_{0}}$. 
Hence, using expression (3), it is possible to extend the classical stiffness model of the serial manipulator by modifying the virtual spring parameters in accordance with the compensator properties. While in the paper this approach has been used for the particular compensator type, similar idea can be evidently ap-plied to other compensators. It is also worth mentioning that the geometrical and elastostatic parameters of gravity compensators ($\alpha$, $a$, $L$ and ${{K}_{c}}$, ${{s}_{0}}$ for the presented case) usually are not included in datasheets. For this reason, these parameters should be identified from the calibration experiments (see Section IV).

\section{Extension of the VJM-based approach}

Stiffness of a serial robot highly depends on its configuration and is defined by the Cartesian stiffness matrix. Using the VJM-based approach adopted in this pa-per, the manipulator can be presented as the sequence of rigid links separated by the actuators and virtual flexible joints incorporating all elastostatic properties of flexible elements [6]. In accordance with these assumptions, the elastostatic model of the serial robot can be presented as shown in Fig. 2

\begin{figure}[hbt]
\center
\includegraphics{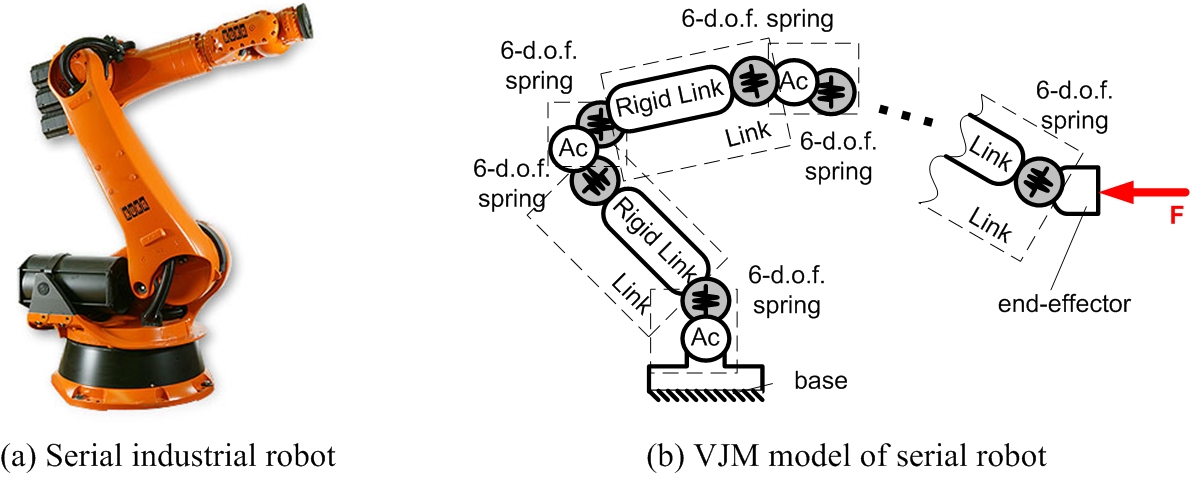}
\caption{Serial industrial robot (a) and its VJM-based stiffness model (b).}
\label{Figure:2}
\end{figure}

For this serial robot, the end-effector location $\mathbf{t}$ can be defined by the vector function $\mathbf{g}\left( \mathbf{q},\mathbf{\theta } \right)$, where 	$\mathbf{q},\,\,\mathbf{\theta }$ denote the vectors of the actuator and virtual joint coordinates respectively. It can be proved that the static equilibrium equations can be written as $\mathbf{J}_{\text{ }\!\!\theta\!\!\text{ }}^{\text{T}}\cdot \mathbf{F}={{\mathbf{K}}_{\text{ }\!\!\theta\!\!\text{ }}}\cdot \mathbf{\theta }$, where $\mathbf{J}_{\text{ }\!\!\theta\!\!\text{ }}^{{}}={\partial \mathbf{g}\left( \mathbf{q},\mathbf{\theta } \right)}/{\partial \mathbf{\theta }}\;$ is the Jacobian matrix, the matrix ${{\mathbf{K}}_{\text{ }\!\!\theta\!\!\text{ }}}=diag\left( {{\mathbf{K}}_{{{\text{ }\!\!\theta\!\!\text{ }}_{1}}}},...,{{\mathbf{K}}_{{{\text{ }\!\!\theta\!\!\text{ }}_{\text{n}}}}} \right)$ aggregates stiffnesses of all virtual springs and $\mathbf{F}$ is the external loading applied to the robot end-effector. In order to find the desired stiffness matrix ${{\mathbf{K}}_{\text{C}}}$, the force-deflection relation should be linearized in the neighborhood of the current configuration $\mathbf{q}$ and presented in the form $\mathbf{F}={{\mathbf{K}}_{\text{C}}}\cdot \Delta \mathbf{t}$, where $\Delta \mathbf{t}$ is the end-effector deflection caused by the external loading $\mathbf{F}$. After relevant transformations, one can get the following expression for the desired stiffness matrix [4]

\begin{equation}\label{Eq:5}
\begin{matrix}
   {{\mathbf{K}}_{\text{C}}}  \\
\end{matrix}={{\left( \mathbf{J}_{\text{ }\!\!\theta\!\!\text{ }}^{{}}\cdot \,\mathbf{K}_{\text{ }\!\!\theta\!\!\text{ }}^{-1}\cdot \,\mathbf{J}_{\text{ }\!\!\theta\!\!\text{ }}^{\text{T}} \right)}^{-1}}
\end{equation}
which depends on the manipulator geometry and its elastostatic properties. This classical expression has been originally derived for the case of manipulators with-out gravity compensators, where the matrix 	${{\mathbf{K}}_{\text{ }\!\!\theta\!\!\text{ }}}$ is constant. 
To integrate into this model the gravity compensator, it is proposed to replace the classical joint stiffness matrix 	$\mathbf{K}_{\text{ }\!\!\theta\!\!\text{ }}^{0}$ (that takes into account elastostatic properties of serial manipulator without gravity compensator) by the sum of two matrices

\begin{equation}\label{Eq:6}
\begin{matrix}
   \mathbf{K}_{\text{ }\!\!\theta\!\!\text{ }}^{{}}  \\
\end{matrix}=\mathbf{K}_{\text{ }\!\!\theta\!\!\text{ }}^{0}+\mathbf{K}_{\text{ }\!\!\theta\!\!\text{ }}^{GC}(\mathbf{q})
\end{equation}
where the second term $\mathbf{K}_{\text{ }\!\!\theta\!\!\text{ }}^{GC}(\mathbf{q})$ is configuration-dependent and takes into account the elastostatic properties of the gravity compensator (by transforming them in the virtual joint space). In practice, the matrix $\mathbf{K}_{\text{ }\!\!\theta\!\!\text{ }}^{GC}(\mathbf{q})$ may include either one or several non-zero elements (depending on the number of compensators). These elements ale located on the matrix diagonal and are computed using (3). 
Hence, by modification of the manipulator joint stiffness matrix in accordance with expression (6), it is possible to integrate the compensator parameters into the classical VJM-based model of the serial manipulator. It should be mentioned that this idea can be also applied for other types of compensators.

\section{Application example}

Let us illustrate the efficiency of the developed stiffness modeling technique by applying it to the compliance error compensation in the robotic-based machining performed by the industrial robot KUKA KR-270. This robot is equipped with the spring-based gravity compensator located between the first and second links (which leads to the influence on the second actuated joint). In accordance with the considered specifications, the technological process should be performed in the square area of the size 2000 mm×2000 mm located at the height 500 mm over the floor level (see Fig, 3 for more details). For comparison purpose, it is assumed that machining force is constant throughout the working area and it is equal to $\mathbf{F}={{(0,360N,560N,0,0,0)}^{T}}$, which corresponds to a typical milling process.

\begin{figure}[hbt]
\center
\includegraphics{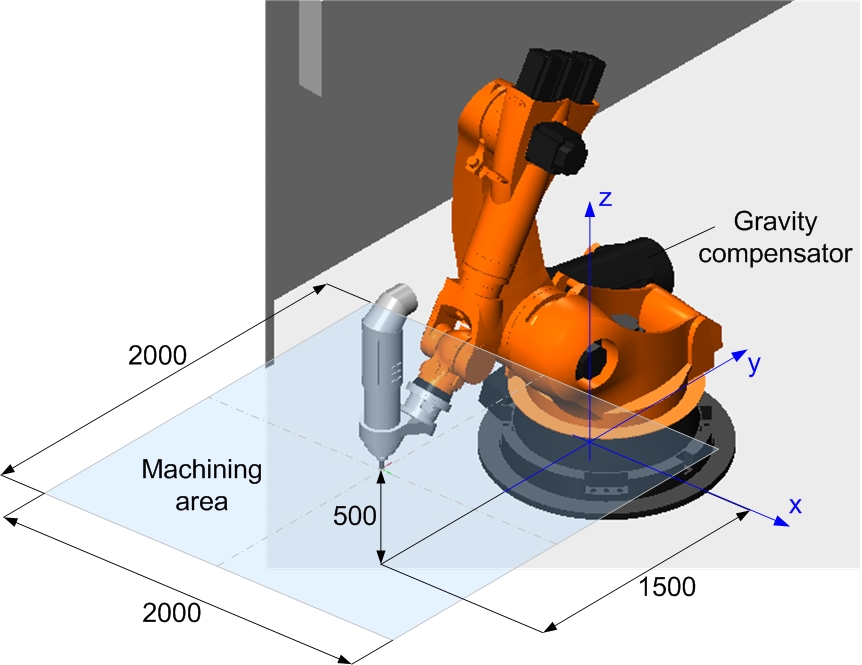}
\caption{Industrial robot Kuka KR270 with required machining area.}
\label{Figure:3}
\end{figure}

To obtain the desired stiffness model, special calibration experiments have been conducted and dedicated identification procedures have been applied. These yield the desired geometrical and elastostatic parameters of the robot with gravity compensator presented in Table 1. From these data, the classical joint stiffness matrix $\mathbf{K}_{\text{ }\!\!\theta\!\!\text{ }}^{0}$ has been constructed straightforwardly (from parameters ${{k}_{1}},...{{k}_{6}}$). To integrate the gravity compensator properties, the matrix $\mathbf{K}_{\text{ }\!\!\theta\!\!\text{ }}^{GC}(\mathbf{q})$ has been computed using expression (3) and values $k_c$, $L$, $a_x$ and $a_y$ for any robot configuration $\mathbf{q}$. Relevant computations have been done throughout the required working area, where the end-effector deflections has been evaluated for given machining force $\mathbf{F}$. The computational results are summarized in Fig. 4, where the end-effector compliance errors are presented. These results show that the compliance errors are not negligible here and vary from 0.34 to 3.5 mm. So, to achieve the desired precision, it is reasonable to apply the error compensation technique, which is based on the reliable stiffness model.

\begin{table}
\caption{Elastostatic and geometrical parameters of robot with gravity compensator.}
	\centering
	\begin{tabular}{p{2.8cm}p{2.8cm}p{2.8cm}p{2.8cm}}
\hline\noalign{\smallskip}
Parameters & Units & Values & CIs \\
\hline\noalign{\smallskip}
$k_1$ & $[rad\cdot m/N]$ & $3.774\cdot 10^{-6}$ & --- \\
$k_2$ & $[rad\cdot m/N]$ & $0.302\cdot 10^{-6}$ & $±0.004\cdot 10^{-6}$ \\
$k_3$ & $[rad\cdot m/N]$ & $0.406\cdot 10^{-6}$ & $±0.008\cdot 10^{-6}$ \\
$k_4$ & $[rad\cdot m/N]$ & $3.002\cdot 10^{-6}$ & $±0.115\cdot 10^{-6}$ \\
$k_5$ & $[rad\cdot m/N]$ & $3.303\cdot 10^{-6}$ & $±0.162\cdot 10^{-6}$ \\
$k_6$ & $[rad\cdot m/N]$ & $2.365\cdot 10^{-6}$ & $±0.095\cdot 10^{-6}$ \\
$k_c$ & $[rad\cdot m/N]$ & $0.144\cdot 10^{-6}$ & $±0.031\cdot 10^{-6}$ \\
$s_0$ & $[mm]$ & $458.00$ & $±27.0 $ \\
$L$ & $[mm]$ & $184.72$ & $±0.06 $ \\
$a_x$ & $[mm]$ & $685.93$ & $±0.70 $ \\
$a_y$ & $[mm]$ & $120.30$ & $±0.69 $ \\

\hline\noalign{\smallskip}
\end{tabular}
	
	\label{Table:1}
\end{table}

\begin{figure}[tb]
\center
\includegraphics{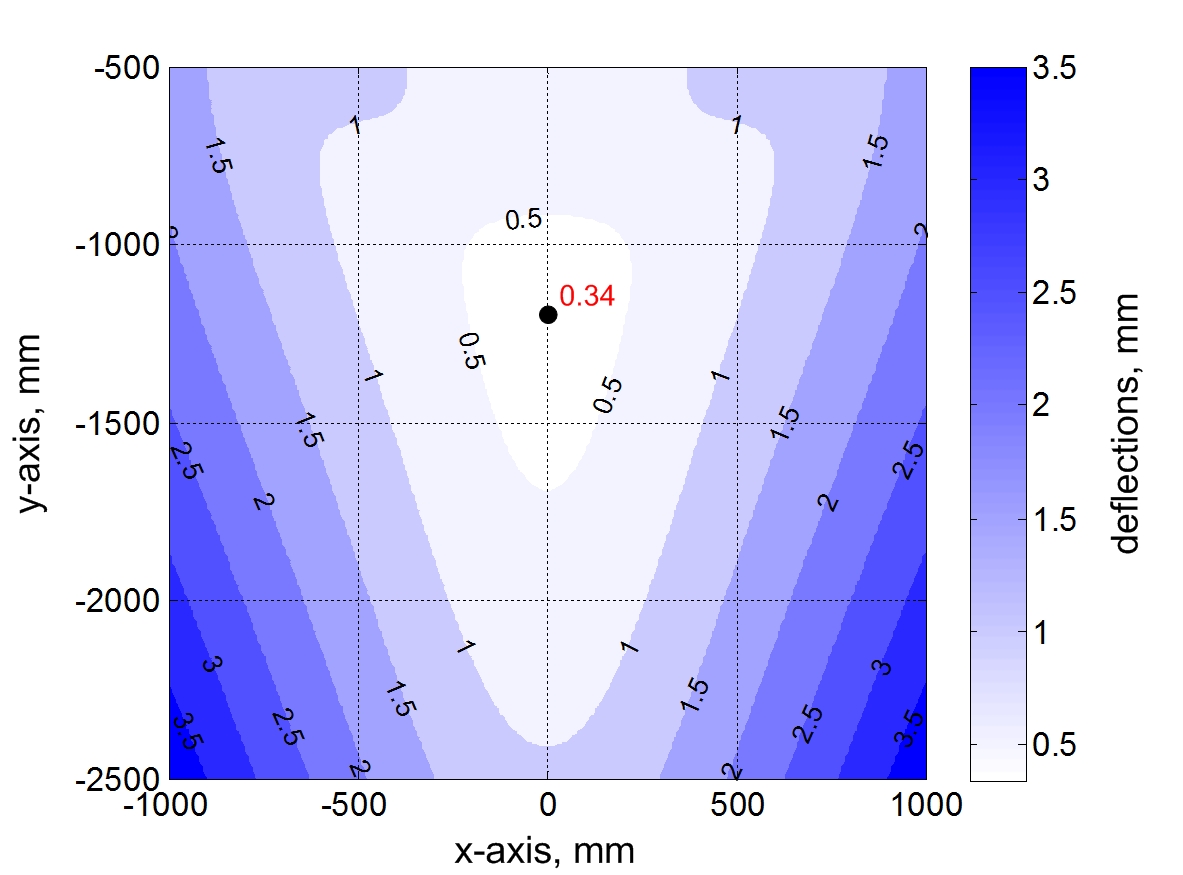}
\caption{The end-effector deflections caused by machining forces throughout working area.}
\label{Figure:4}
\end{figure}

For comparison purposes, two alternative stiffness models have been examined. The first one is based on the classical assumptions and takes into account the stiffness properties of the actuated joints only (without gravity compensator). The second model integrates the gravity compensator in accordance with the approach proposed in this paper. Both the models have been used for the compliance error compensation for the described above manufacturing task. Relevant results are presented in Fig. 5, which shows the difference in the compliance error compensation while applying the classical and the proposed approaches. As follows from the figure, ignoring the gravity compensator influence may lead to the position error over/under compensation of the order of 0.07 mm, which is not admissible for the manufacturing processes employed in the aerospace applications studied here. Hence, to ensure the high precision for the robotic-based machining, the compliance error compensation technique must rely on the stiffness model, which takes into account the impact of the gravity compensator, in accordance with the approach developed in this paper.

\begin{figure}[tb]
\center
\includegraphics{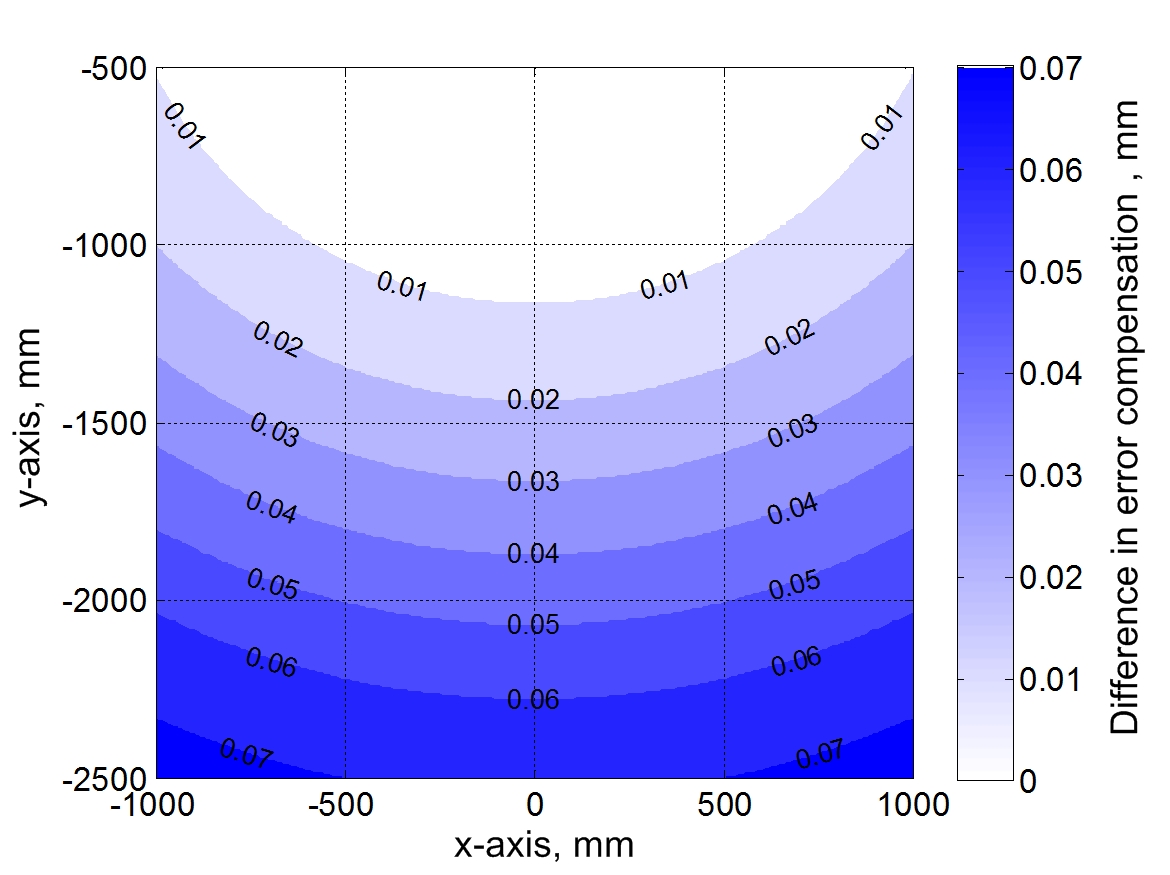}
\caption{Difference in stiffness error compensation between two strategies.}
\label{Figure:5}
\end{figure}

\section{Conclusions}
The paper presents a new approach for the stiffness modeling of robots with the spring-based gravity compensators, which are located between the manipulator sequential links. Using this approach, the compensator has been replaced by an equivalent non-linear virtual spring integrated in the corresponding actuated joint. This methodology allowed us to extend the VJM-based modeling technique for the case of manipulators with closed-loops induced by the gravity compensators via using configuration-dependent joint stiffness matrix. Efficiency of the developed approach and its industrial value have been confirmed by an application ex-ample, which deals with robotic-based milling of large-dimensional parts for aerospace industry. Feature work will deal with integration of this modeling approach into the robotic software CAD system.

\begin{acknowledgement}
The work presented in this paper was partially funded by the ANR, France (Project ANR-2010-SEGI-003-02-COROUSSO) and Project ANR ROBOTEX. The authors also thank Fabien Truchet, Guillaume Gallot, Joachim Marais and Sébastien Garnier for their great help with the experiments.\end{acknowledgement}

\end{document}